\documentclass[a4paper]{spie}

\usepackage{amsmath,amsfonts,amssymb}
\usepackage{graphicx}
\usepackage{booktabs}
\usepackage{array}
\usepackage{url}
\usepackage[T1]{fontenc}
\usepackage{microtype}
\usepackage{xcolor}
\usepackage{placeins}
\usepackage{float}
\usepackage{indentfirst}
\usepackage[colorlinks=true,linkcolor=blue,citecolor=blue,urlcolor=blue]{hyperref}

\title{SRE-FER: Regional residual evidence learning for mitigating local evidence dilution in fine-grained facial expression recognition}

\author{Jiaye Song\textsuperscript{*,\textdagger}}
\author{Ruochen Zhang\textsuperscript{*}}
\author{Yuliang Wang\textsuperscript{*}}
\author{Jiaqi Wu}
\affil{Tiangong University, No. 399 Binshuixi Road, Xiqing District, Tianjin 300387, China}
\authorinfo{\textsuperscript{*}These authors contributed equally. \textsuperscript{\textdagger}Corresponding author: Jiaye Song; E-mail: a17778032691@163.com\\Ruochen Zhang; E-mail: qianyuanshi26@163.com; Yuliang Wang; E-mail: wangyuliang@tiangong.edu.cn\\Jiaqi Wu; E-mail: 2311640202@tiangong.edu.cn}

\begin{document}

\maketitle
\enlargethispage{-2\baselineskip}

\begin{abstract}
Fine-grained facial expression recognition (FER) hinges on capturing subtle muscular cues that distinguish adjacent emotions. Yet capturing these cues presents a dilemma. Detector-based methods depend on fragile landmark pipelines, whereas we find that directly transferring foundation models such as DINOv3 under conventional global readouts can cause \textbf{local evidence dilution}: early global aggregation washes out sparse muscular signals and leaves persistent confusion between categories such as \textit{fear}/\textit{surprise} and \textit{sad}/\textit{neutral}. To recover this evidence, we propose SRE-FER, a readout-level \textbf{regional residual evidence} learning framework. Its core module, RERA, adds zero-initialized residual logits that refine class boundaries while preserving the backbone's global prediction. Training-time action unit (AU) guidance steers regional features toward expression-relevant areas using Facial Action Coding System (FACS)-based anatomical priors, without requiring an external facial pipeline at inference. An optional Full setting further routes sample-specific non-redundant tokens. On three benchmarks, SRE-FER attains 92.76\% on RAF-DB, 91.32\% on FERPlus, and 67.78\% on AffectNet-7, demonstrating highly competitive performance compared to existing FER methods.
\end{abstract}

\quad\textbf{Conference Topic:} Computer Vision and Robot Vision

\section{Introduction}
\label{sec:intro}

\begin{figure}[ht]
\centering
\includegraphics[width=0.98\textwidth]{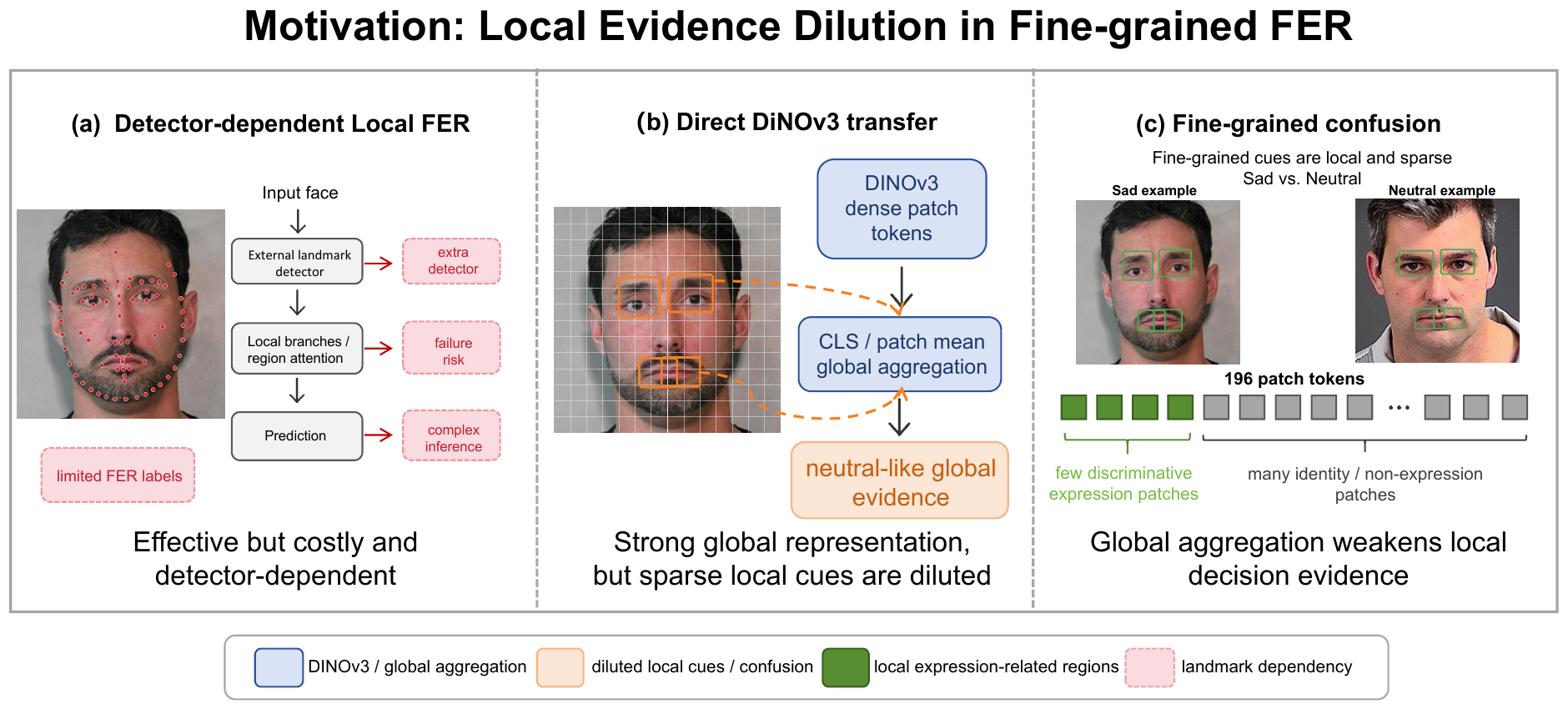}
\caption{Motivation of local evidence dilution in fine-grained FER.}
\label{fig:intro}
\end{figure}

It has been a long-standing goal in facial expression recognition (FER) to learn representations that are both robust to variations in identity, pose, and illumination and sensitive to subtle muscular changes. The balance is difficult because adjacent expressions often differ only in small facial areas, such as the eyebrow and eyelid cues shared by \textit{fear}/\textit{surprise} or the mouth-corner cues separating \textit{sad}/\textit{neutral}. Existing FER methods mainly follow two routes. Holistic methods strengthen image-level classifiers through robust objectives and transformer modeling~\cite{ref_scn,ref_dacl,ref_eac,ref_transfer}, but may miss sparse muscular cues. Region-focused methods expose these cues through attention masks, landmarks, or facial-structure branches~\cite{ref_poster,ref_lan}, at the cost of additional task-specific machinery. This creates the two-sided tension illustrated in Fig.~\ref{fig:intro}: one route is simple but often insufficiently detailed, whereas the other is targeted but less self-contained.

This tension motivates us to revisit self-supervised vision foundation models. Representative FER systems still begin from supervised CNN or face-recognition encoders, such as ResNet-18 in SCN and IR-50 in POSTER~\cite{ref_scn,ref_poster}. DINOv3 instead learns without expression labels and provides dense patch tokens with strong spatial correspondence~\cite{ref_dinov3}, making it a natural candidate for detector-free cue modeling. Rather than treating it as a plug-and-play backbone, we study how these dense tokens behave under fine-grained FER readouts. However, conventional class-token (CLS) and mean-pooled readouts aggregate patches early, allowing sparse muscular cues to be absorbed into a holistic facial representation. We call this readout-level failure \textbf{local evidence dilution}. The key question is how to recover these cues from DINOv3 patch tokens while retaining the stability of the global classifier.

We therefore present \textbf{SRE-FER}, a readout-level framework that treats regional cues as residual evidence for the global prediction. Its Regional Evidence Residual Adapter (RERA) partitions DINOv3 patch tokens into spatial regions, pools region-specific evidence, and produces zero-initialized residual logits. Training-time action unit (AU) guidance introduces Facial Action Coding System (FACS)-based anatomical priors, but is removed at inference; SRE-FER therefore requires no test-time landmark detector or external facial-structure pipeline. In the Full setting, a non-redundant evidence branch further routes complementary sample-specific cues.

The paper makes three contributions:
\begin{enumerate}
    \item \textbf{Analytically}, we identify local evidence dilution in conventional readouts of dense DINOv3 tokens, where early aggregation weakens sparse expression cues.
    \item \textbf{Methodologically}, RERA adds zero-initialized regional residual logits, while training-time AU guidance provides anatomical regularization without an inference-time facial pipeline.
    \item \textbf{Experimentally}, SRE-FER demonstrates highly competitive performance on three benchmarks; progressive ablations and class-level diagnostics trace the gains.
\end{enumerate}

\section{Related Work}

\subsection{Holistic and Region-Based Modeling for Fine-Grained FER}

Holistic FER methods improve label robustness through uncertainty-aware objectives and attentive losses~\cite{ref_scn,ref_dacl}, while relation-aware transformers strengthen image-level representations~\cite{ref_eac,ref_transfer}. Region-focused methods retain more spatial structure: RAN emphasizes pose-robust local areas~\cite{ref_ran}, LAN learns local attention masks~\cite{ref_lan}, and POSTER combines image and landmark streams~\cite{ref_poster}. Ambiguity modeling and multi-scale fusion offer complementary improvements~\cite{ref_dmue,ref_vtff}. Together, these studies establish the value of local evidence, but they either compress it into an image-level representation or introduce a task-specific local branch. SRE-FER instead reads regional evidence from patch tokens already produced by the backbone.

\subsection{Foundation Model Adaptation and Local Evidence Dilution}

Foundation-model FER has largely followed a semantic route. DFER-CLIP and EmoCLIP adapt vision-language representations to expression recognition~\cite{ref_dfer_clip,ref_emoclipl}, while MPA-FER aligns global and local visual features with multimodal prompts~\cite{ref_mpafer}. LoRA and visual prompt tuning provide general mechanisms for transferring large pre-trained models with few task-specific parameters~\cite{ref_lora,ref_vpt}. A recent norm-referenced encoding study also tested a final DINOv3 CLS feature on the synthetic FERG dataset~\cite{ref_mdnre}. It establishes the compatibility of the backbone with FER, but leaves a visual-side question open: how should dense self-supervised tokens be read on in-the-wild benchmarks?

DINOv3 offers a useful setting for this question: its patch tokens preserve spatial detail, whereas the usual CLS or mean-pooled readout remains global. SRE-FER acts at this interface. It leaves the global path intact and adds regional correction at the logits, allowing dense visual evidence to influence the decision without introducing a language-alignment or detector pipeline.

\subsection{AU Priors and Structured Constraints}

The Facial Action Coding System (FACS)~\cite{ref_facs} defines anatomically grounded action units used to characterize expression-related facial movements. Tools such as OpenFace~\cite{ref_openface} enable automatic AU detection; JAA-Net~\cite{ref_jaanet} jointly models AU detection and face alignment. Prior AU- or landmark-guided pipelines often rely on external facial structure estimators during inference, adding computational cost and risking failure under occlusion~\cite{ref_poster,ref_mediapipe}. SRE-FER bridges global foundation-model readouts and detector-dependent regional modeling by using AU-derived guidance only during training and reading regional evidence directly from patch tokens at inference.

\section{Method}

\subsection{Overall Framework}

SRE-FER is built around a simple principle: retain the global prediction and introduce local evidence as a residual correction (Fig.~\ref{fig:arch}). A partially frozen DINOv3 encoder maps image $x$ to a CLS token $\mathbf{h}_{\mathrm{cls}}$ and $N$ spatial patch tokens $\mathbf{P}$:

\begin{equation}
[\mathbf{h}_{\mathrm{cls}}, \mathbf{P}] = F_{\theta}(x), \quad \mathbf{P} = [\mathbf{p}_1, \ldots, \mathbf{p}_N] \in \mathbb{R}^{N \times D}
\end{equation}

A patch-aware global path combines $\mathbf{h}_{\mathrm{cls}}$ with the patch mean to form base logits $\mathbf{o}_g$. In parallel, RERA pools $K$ spatial regions and produces residual logits $\mathbf{o}_r$:

\begin{equation}
\mathbf{o} = \mathbf{o}_g + \gamma \mathbf{o}_r, \quad p(y=c \mid x) = \frac{\exp(o_c)}{\sum_{j=1}^{C} \exp(o_j)}
\end{equation}

Here, $\gamma$ is a learnable scalar initialized to zero. Consequently, the initial joint logits and predictions are exactly those of the global classifier. Unlike an unconstrained concatenation or late-fusion head, which reparameterizes the decision from the outset, RERA leaves $\mathbf{o}_g$ explicit and learns an additive correction to its class margins. Deploy. contains only the encoder, global readout, and RERA; Full additionally routes tokens that are poorly represented by the dominant patch basis.

\subsection{Dense Token Encoding and Base Readout}

\begin{figure}
\centering
\includegraphics[width=0.85\textwidth]{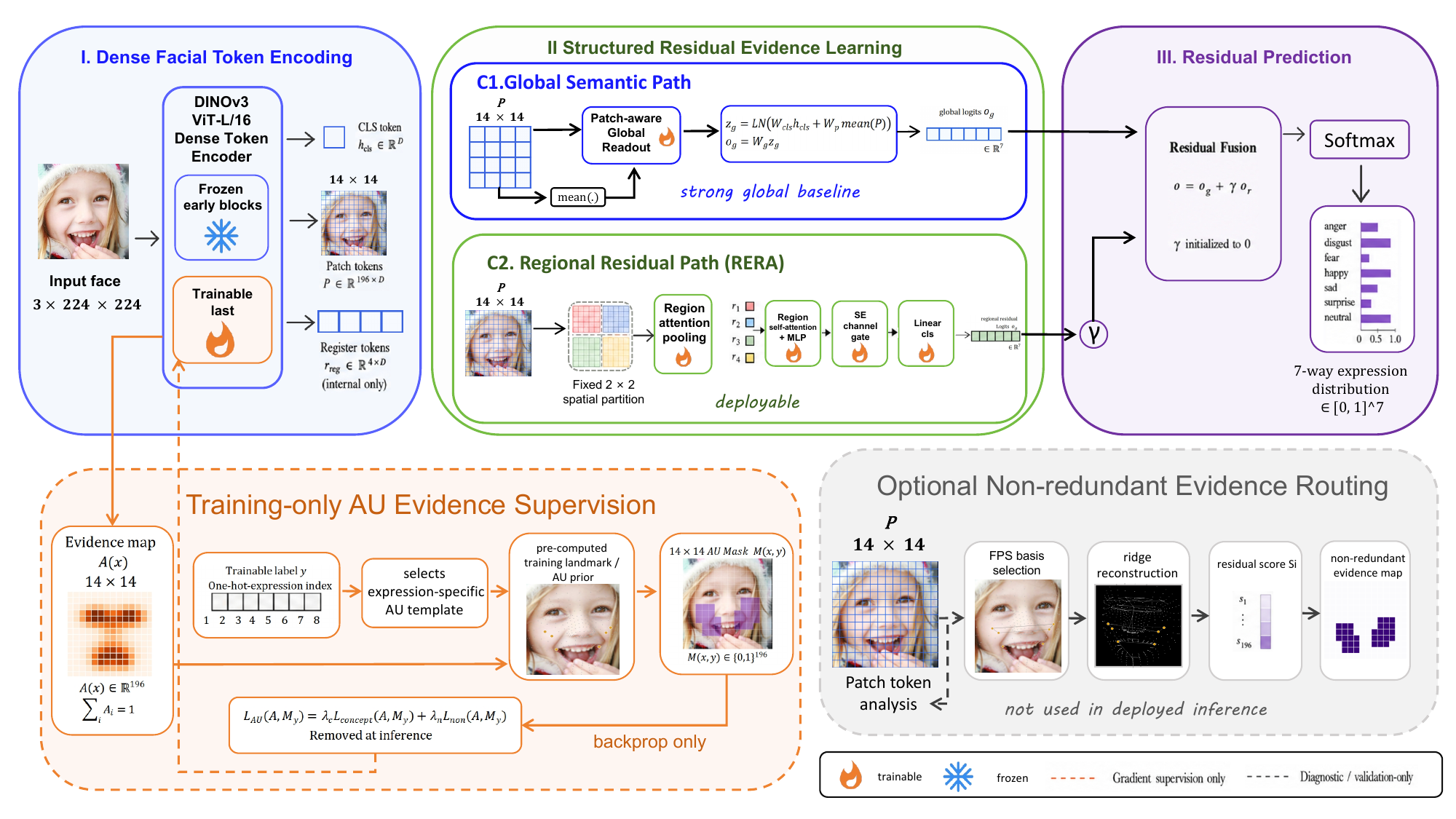}
\caption{Core Deploy. architecture of SRE-FER. The DINOv3 global path remains intact, while RERA reads regional patch evidence and supplies a zero-initialized residual correction. Training-only AU guidance and optional Full routing are detailed in Secs.~\ref{sec:au} and~\ref{sec:routing}.}
\label{fig:arch}
\end{figure}

With a $224 \times 224$ input, DINOv3 ViT-L/16 yields a $14 \times 14$ patch grid ($N = 196$, $D = 1024$; Fig.~\ref{fig:arch}). Register tokens are discarded; the first 20 Transformer blocks are frozen, while the last 4 blocks and final LayerNorm are fine-tuned. We retain the spatial tokens as the local feature basis for subtle muscular changes.

The patch-aware readout jointly exploits the CLS token and patch statistics to produce stable base logits $\mathbf{o}_g$, where LN denotes layer normalization:

\begin{equation}
\bar{\mathbf{p}} = \frac{1}{N} \sum_{i=1}^{N} \mathbf{p}_i,\qquad
\mathbf{z}_g = \mathrm{LN}\left(W_c \mathbf{h}_{\mathrm{cls}} + W_p \bar{\mathbf{p}}\right),\qquad
\mathbf{o}_g = W_g \mathbf{z}_g
\end{equation}

This readout preserves CLS-level semantics while exposing patch statistics to the subsequent regional correction.

\subsection{AU Prior-Guided Evidence Constraint}\label{sec:au}

We introduce a training-only AU-guided constraint that encourages anatomically plausible attention distributions.

\subsubsection{Sample-Level Prior Mask} MediaPipe~\cite{ref_mediapipe} extracts 478 normalized landmarks during offline training preprocessing. We use the fixed FACS-derived mapping in Table~\ref{tab:au_prior} to define AU-relevant landmarks for each expression. A point $(u,v)$ is assigned to patch $(\lfloor14u\rfloor,\lfloor14v\rfloor)$ after clipping both indices to $[0,13]$. Landmark and AU overlaps are merged by binary union; sparse class masks are expanded by at most one Chebyshev-neighborhood ring. Image and mask are horizontally flipped together. The result is a sample-specific mask $M(x,y)\in\{0,1\}^N$, with relevant and irrelevant index sets $\mathcal{S}_y$ and $\bar{\mathcal{S}}_y$.

\begin{table}[!htb]
\caption{AU priors used to construct training-time masks.}\label{tab:au_prior}
\centering
\scriptsize
\setlength{\tabcolsep}{1.2pt}
\begin{tabular}{@{}lll@{\hspace{0.5em}}lll@{}}
\toprule
\textbf{Expr.} & \textbf{AUs} & \textbf{Regions} & \textbf{Expr.} & \textbf{AUs} & \textbf{Regions} \\
\midrule
Surprise & 1/2/5/26 & brow, eyelid, jaw/mouth & Disgust & 9/10 & nose, upper lip \\
Fear & 1/2/4/5/20/26 & brow, eyelid, mouth, jaw & Happy & 6/12 & cheek, mouth corners \\
Sad & 1/4/15 & inner brow, mouth corners & Anger & 4/5/7/23 & brow, eyelid, lip \\
Neutral & -- & positive mask disabled & & & \\
\bottomrule
\end{tabular}
\end{table}

\subsubsection{Constraint Loss} Let $\mathbf{A}(x) \in \mathbb{R}^N$ denote the CLS-to-patch attention from the last Transformer layer, obtained by averaging attention weights across all heads, excluding register tokens, and renormalizing to sum to 1. We define positive and negative patch-level constraints:

\begin{equation}
\mathcal{L}_{\mathrm{concept}} = -\frac{1}{|\mathcal{S}_y|} \sum_{i \in \mathcal{S}_y} \log(A_i(x) + \epsilon),\quad
\mathcal{L}_{\mathrm{non}} = -\frac{1}{|\bar{\mathcal{S}}_y|} \sum_{i \in \bar{\mathcal{S}}_y} \log(1 - A_i(x) + \epsilon)
\end{equation}

The total AU loss is $\mathcal{L}_{\mathrm{AU}} = \lambda_{\mathrm{pos}} \mathcal{L}_{\mathrm{concept}} + \lambda_{\mathrm{neg}} \mathcal{L}_{\mathrm{non}}$. For \textit{neutral}, $\mathcal{L}_{\mathrm{concept}}$ is disabled and $\mathcal{L}_{\mathrm{non}}$ penalizes attention on the union of expression-related regions. A 1,400-image RAF-DB training probe (200 per class) detected 1,293 faces (92.4\%); rates ranged from 82.5\% for \textit{anger} to 96.5\% for \textit{surprise} and \textit{happy}. Failed detections produce an all-zero mask and zero weight for both AU terms, so ordinary classification supervision remains active. AU guidance is removed entirely at inference.

\subsection{RERA}

RERA (green modules, Fig.~\ref{fig:arch}) extracts incremental regional evidence from patch tokens and corrects decision boundaries with residual logits, without replacing the base classifier.

\subsubsection{Regional Partitioning and Evidence Pooling} The $14 \times 14$ grid is divided into $K = 4$ non-overlapping $7 \times 7$ quadrants $\{\mathcal{R}_k\}_{k=1}^K$. This is a coarse coordinate frame for the aligned-face inputs used here, not a claim of optimal semantic segmentation. It gives every region 49 candidate tokens without a per-image detector; a learned query $\mathbf{q}_k \in \mathbb{R}^D$ then selects evidence within each quadrant:

\begin{equation}
\alpha_{k,i} = \frac{\exp(\mathbf{q}_k^\top \mathbf{p}_i / \sqrt{D})}{\sum_{j \in \mathcal{R}_k} \exp(\mathbf{q}_k^\top \mathbf{p}_j / \sqrt{D})}, \quad \mathbf{r}_k = \sum_{i \in \mathcal{R}_k} \alpha_{k,i} \mathbf{p}_i
\end{equation}

This yields regional tokens $\mathbf{R} = [\mathbf{r}_1, \ldots, \mathbf{r}_K] \in \mathbb{R}^{K \times D}$. Within-region attention avoids uniform averaging, while the following cross-region block allows information to cross the hard quadrant boundaries.

\subsubsection{Cross-Region Coordination} Lightweight multi-head attention (MHA), followed by average pooling and squeeze-and-excitation (SE) gating, models inter-region interactions. The multilayer perceptron (MLP) provides a token-wise update:

\begin{equation}
\widetilde{\mathbf{R}} = \mathbf{R} + \mathrm{MHA}(\mathrm{LN}(\mathbf{R})), \quad \widehat{\mathbf{R}} = \widetilde{\mathbf{R}} + \mathrm{MLP}(\mathrm{LN}(\widetilde{\mathbf{R}})), \quad \mathbf{z}_r = \mathrm{SE}\left(\frac{1}{K} \sum_{k=1}^{K} \widehat{\mathbf{r}}_k\right)
\end{equation}

The residual logits are $\mathbf{o}_r = W_r \mathbf{z}_r$. For ground-truth class $y$ and confusion class $c$, the margin after RERA correction is:

\begin{equation}
m_{y,c} = (\mathbf{o}_{g,y} - \mathbf{o}_{g,c}) + \gamma(\mathbf{o}_{r,y} - \mathbf{o}_{r,c})
\end{equation}

RERA thus learns a correction term rather than an independent replacement classifier: the base margin remains explicit, while $\gamma$ controls the regional contribution. Direct regional supervision (Sec.~\ref{sec:train_obj}) trains $\mathbf{o}_r$ even while the joint path initially equals the base classifier.

\subsection{Non-Redundant Evidence Routing}\label{sec:routing}

AU priors mark plausible anatomy but not sample-specific increments. Full therefore uses farthest point sampling (FPS) and ridge reconstruction as an optional routing test, separate from the core Deploy. claim. Given $\mathbf{P} \in \mathbb{R}^{N \times D}$, FPS selects $B$ bases, ridge regression reconstructs $\widehat{\mathbf{p}}_i$, and $s_i=\|\mathbf{p}_i-\widehat{\mathbf{p}}_i\|_2$ scores evidence outside the dominant basis. The top-$k$ tokens are pooled and fused with RERA (Stage~G); Deploy. (Stage~F) omits this branch.

\subsection{Training Objective and Stabilization}\label{sec:train_obj}

For RAF-DB, $\mathcal{L}_{\mathrm{main}}$ is cross-entropy on the joint output $\mathbf{o}$, with auxiliary regional supervision $\mathcal{L}_{\mathrm{region}}$ (cross-entropy on $\mathbf{o}_r$) and base supervision $\mathcal{L}_{\mathrm{cls}}$ (cross-entropy on $\mathbf{o}_g$):

\begin{equation}
\mathcal{L} = \mathcal{L}_{\mathrm{main}} + \lambda_{\mathrm{AU}} \mathcal{L}_{\mathrm{AU}} + \lambda_r \mathcal{L}_{\mathrm{region}} + \lambda_{\mathrm{cls}} \mathcal{L}_{\mathrm{cls}}
\end{equation}

Although $\gamma$ is initialized to zero, the residual branch receives direct supervision through $\mathcal{L}_{\mathrm{region}}$, so $\mathbf{o}_r$ learns meaningful corrections before being injected into the joint output.

For AffectNet-7, $\mathcal{L}_{\mathrm{main}}$ is standard cross-entropy as in the RAF-DB setup. For FERPlus, $\mathcal{L}_{\mathrm{main}}$ is replaced with soft-label cross-entropy. To reduce training variance on small FER datasets, we apply SAM ($\rho = 0.05$)~\cite{ref_sam}, EMA ($\beta=0.999$), and SWA (checkpoint averaging from epoch 8). For stability diagnostics, we optionally use horizontal-flip test-time augmentation (TTA): $\mathbf{o}_{\mathrm{tta}} = \frac{1}{2}(\mathbf{o}(x) + \mathbf{o}(\mathrm{flip}(x)))$.

\section{Experiments}

We evaluate SRE-FER against representative FER methods, then use progressive ablations and fine-grained analyses to examine where the gains appear and how the Deploy. and Full paths differ.

\subsection{Datasets and Evaluation Metrics}

RAF-DB contains seven expression categories with 12,271 training and 3,068 test images. We follow the official split, reserve 10\% of training samples for validation, and report multi-seed stability separately in Sec.~\ref{sec:stability}.

For FERPlus, Table~\ref{tab:datasets} records the standard 8-class protocol (neutral, happy, surprise, sad, anger, disgust, fear, contempt). Following~\cite{ref_poster,ref_transfer}, images with unknown or NF majority votes are excluded, and the remaining votes are normalized as soft labels. AffectNet-7 follows its official seven-class train/validation split. All inputs are resized to $224\times224$. We report Accuracy and additionally use Macro-F1 to examine class behavior on imbalanced RAF-DB.

\begin{table}[!ht]
\caption{Experimental dataset configurations.}\label{tab:datasets}
\centering
\begin{tabular}{lcccc}
\toprule
\textbf{Dataset} & \textbf{Classes} & \textbf{Train} & \textbf{Test/Val} & \textbf{Input} \\
\midrule
RAF-DB & 7 & 12,271 & 3,068 & aligned face, $224\times224$ \\
FERPlus & 8 & standard split & standard split & soft vote labels, $224\times224$ \\
AffectNet-7 & 7 & official train & official val & $224\times224$ \\
\bottomrule
\end{tabular}
\end{table}

\subsection{Implementation Details}\label{sec:impl}

Training runs for 150 epochs with batch size 64 under bf16 precision. We use AdamW with weight decay 0.05; the backbone and task heads use base learning rates of $1 \times 10^{-5}$ and $1 \times 10^{-4}$, respectively, under a cosine schedule. Augmentation combines horizontal flipping, RandAugment, and Random Erasing. The AU loss uses $\lambda_{\mathrm{pos}} = 0.5$ and $\lambda_{\mathrm{neg}} = 1.2$, the RERA auxiliary weight is $\lambda_r = 0.075$, and the base supervision weight is $\lambda_{\mathrm{cls}} = 0.2$. SAM ($\rho = 0.05$), EMA, and SWA from epoch 8 stabilize training. Main evaluation uses no TTA; \emph{stabilized deployment inference} denotes EMA/SWA-smoothed prediction without a test-time AU module, landmark detector, or evidence routing. The Full branch uses FPS $B = 32$, ridge $\alpha = 0.1$, top-$k = 16$, and a learned fusion weight. All experiments run on a single RTX 4090 ($\approx$11.7~GB).

\subsection{Comparison with Representative Methods}

Table~\ref{tab:comparison} reports the comparison under two inference settings. SRE-FER (Deploy.) uses only DINOv3, the base readout, and RERA, achieving 92.47\% on RAF-DB, 91.07\% on FERPlus, and 67.59\% on AffectNet-7. SRE-FER (Full) activates non-redundant evidence routing and further improves the results to 92.76\%, 91.32\%, and 67.78\%, respectively.

\begin{table}[!htb]
\caption{Comparison with representative visual-only FER methods. ``--'' denotes unreported results, and bold marks the column maximum among the listed methods. Deploy. and Full denote the deployment and evidence-routing settings.}\label{tab:comparison}
\centering
\footnotesize
\setlength{\tabcolsep}{3.5pt}
\begin{tabular}{lcccc}
\toprule
\textbf{Method} & \textbf{Venue} & \textbf{RAF-DB} & \textbf{FERPlus} & \textbf{AffectNet-7} \\
\midrule
SCN~\cite{ref_scn} & CVPR'20 & 88.14 & 89.35 & -- \\
DMUE~\cite{ref_dmue} & CVPR'21 & 89.42 & -- & 63.11 \\
TransFER~\cite{ref_transfer} & ICCV'21 & 90.91 & 90.83 & 66.23 \\
EAC~\cite{ref_eac} & ECCV'22 & 90.35 & 89.64 & 65.32 \\
POSTER~\cite{ref_poster} & ICCVW'23 & 92.05 & \textbf{91.62} & 67.31 \\
GLCL-Net~\cite{ref_glclnet} & CAIBDA'24 & 90.71 & 90.79 & 67.17 \\
MFER~\cite{ref_mfer} & TAFFC'25 & 92.08 & 91.09 & 67.06 \\
AU-ViT~\cite{ref_auvit} & TMM'25 & 91.10 & 90.15 & 65.71 \\
WGGLFA~\cite{ref_wgglfa} & Biomimet.'25 & 90.32 & 91.24 & -- \\
\midrule
SRE-FER (Deploy.) & Ours & 92.47 & 91.07 & 67.59 \\
SRE-FER (Full) & Ours & \textbf{92.76} & 91.32 & \textbf{67.78} \\
\bottomrule
\end{tabular}
\end{table}

Across the three benchmarks in Table~\ref{tab:comparison}, SRE-FER (Full) demonstrates highly competitive performance; routing adds 0.29, 0.25, and 0.19 points over Deploy., respectively. MPA-FER is discussed separately because it instead uses frozen CLIP, large-language-model-derived text prompts, and cross-modal alignment~\cite{ref_mpafer}. A progressive DINOv3 ablation is reported in Table~\ref{tab:ablation}.

\subsection{Ablation Study}

Table~\ref{tab:ablation} traces the model from a CLS-only DINOv3 classifier to Full while keeping the backbone, input resolution, and data split fixed. The stages add patch statistics, AU guidance, RERA, SAM, EMA/SWA-based selection, and non-redundant routing; Sec.~\ref{sec:fine_grained} examines the corresponding class-wise predictions.

\begin{table}[!ht]
\caption{Progressive ablation study on RAF-DB.}\label{tab:ablation}
\centering
\small
\begin{tabular}{clccr}
\toprule
\textbf{Stage} & \textbf{Variant} & \textbf{Key Design} & \textbf{Acc (\%)} & \textbf{Gain} \\
\midrule
A & CLS only & semantic token & 90.25 & -- \\
B & CLS + patch mean & dense token statistics & 90.61 & +0.36 \\
C & + AU constraint & anatomy-guided evidence & 90.88 & +0.27 \\
D & + RERA & regional residual logits & 91.80 & +0.92 \\
E & + SAM & sharpness-aware training & 92.11 & +0.31 \\
F & SRE-FER (Deploy.) & + EMA/SWA-based selection & 92.47 & +0.36 \\
G & SRE-FER (Full) & + non-redundant evidence & 92.76 & +0.29 \\
\bottomrule
\end{tabular}
\end{table}

Stages A--D keep DINOv3, the split, and the training budget fixed: patch statistics and AU guidance add 0.36 and 0.27 points, followed by +0.92 from the complete RERA unit. The table does not isolate zero initialization or compare untested concatenation and late-fusion heads. SAM and EMA/SWA-based selection then reach 92.47\%, while optional routing adds 0.29 points.

\subsection{Fine-Grained Category Decoupling and Confusion Matrix Analysis}\label{sec:fine_grained}

\begin{figure}[!tbp]
\centering
\includegraphics[width=0.82\textwidth]{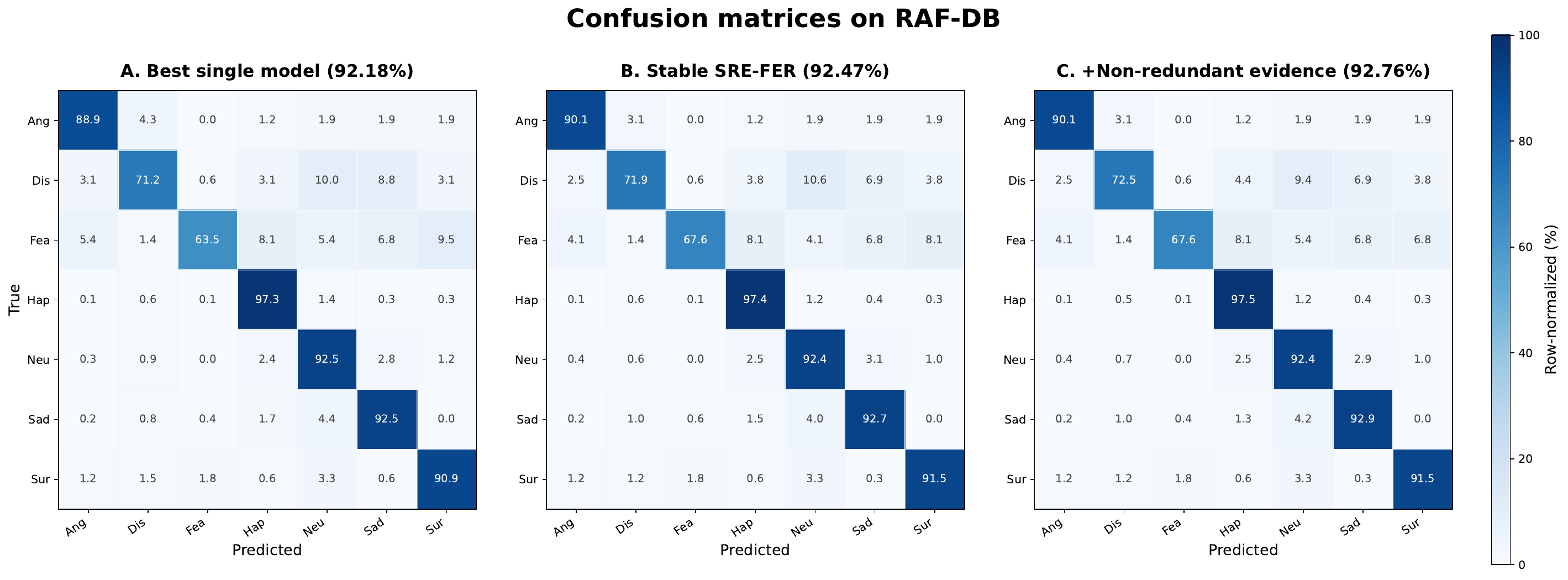}
\caption{RAF-DB confusion matrices for three checkpoints: (A) best single-checkpoint Deploy. without EMA/SWA (92.18\%), (B) SRE-FER Deploy. (92.47\%), and (C) SRE-FER Full (92.76\%). The letters do not denote ablation stages; Table~\ref{tab:repair} uses a separate common pre-routing set.}
\label{fig:cm_triptych}
\end{figure}

If global readout underuses local cues, the largest changes should occur in \textbf{boundary-sensitive classes} rather than only in dominant categories. We therefore analyze grouped F1 (Table~\ref{tab:fine_grained}) and confusion matrix evolution (Fig.~\ref{fig:cm_triptych}) on RAF-DB, where overall accuracy is influenced by \textit{happy} and \textit{neutral}. SRE-FER raises the boundary-sensitive set (\textit{fear}, \textit{disgust}, \textit{sad}, \textit{surprise}) from 79.9\% to 83.5\%; the minority \textit{fear}/\textit{disgust} group shows the largest change.

\begin{figure}[!tbp]
\centering
\begin{minipage}[c]{0.47\textwidth}
\centering
\includegraphics[width=\linewidth]{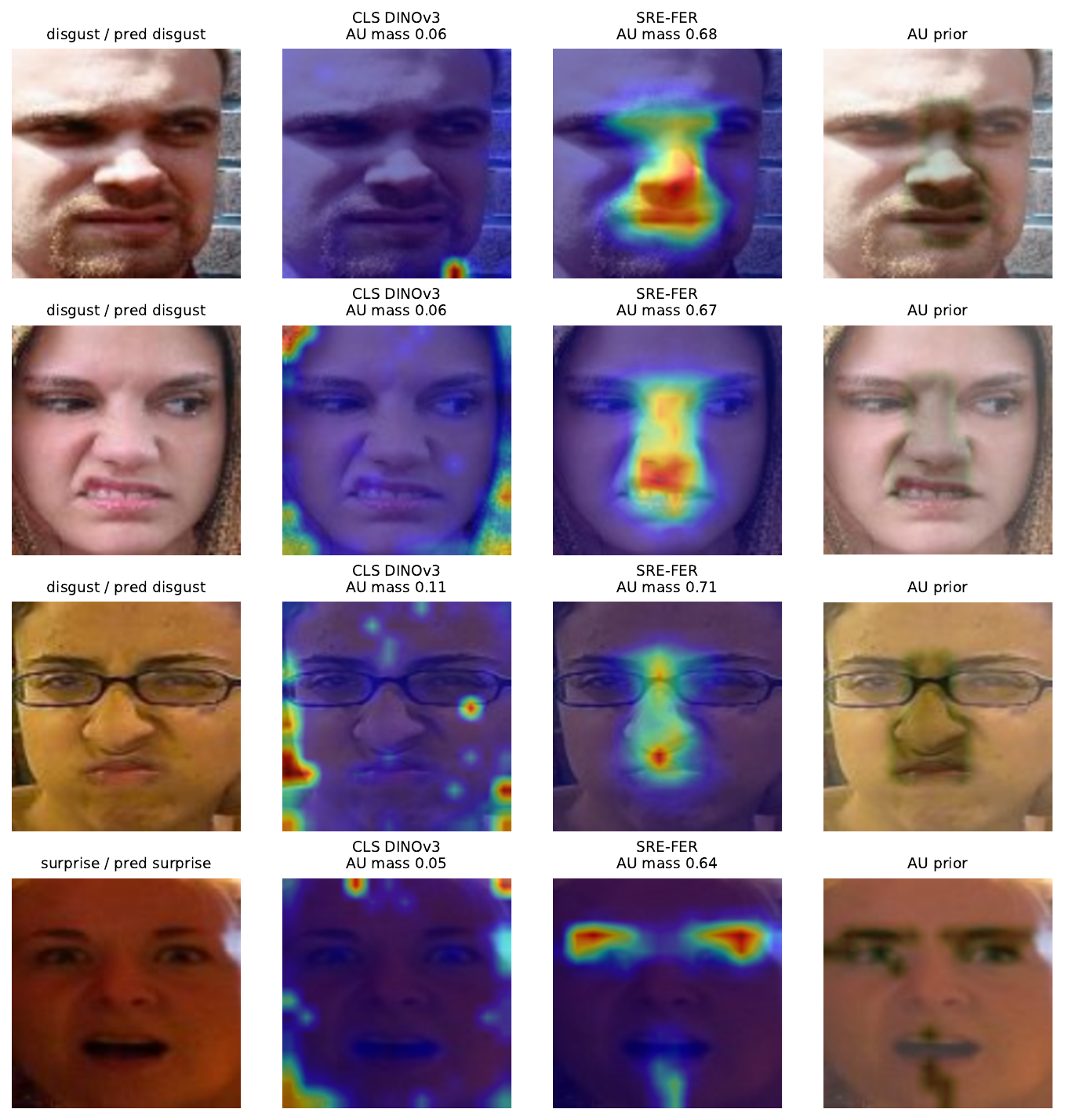}
\par\smallskip\textbf{(a)} Attention evidence
\end{minipage}\hfill
\begin{minipage}[c]{0.47\textwidth}
\centering
\includegraphics[width=\linewidth]{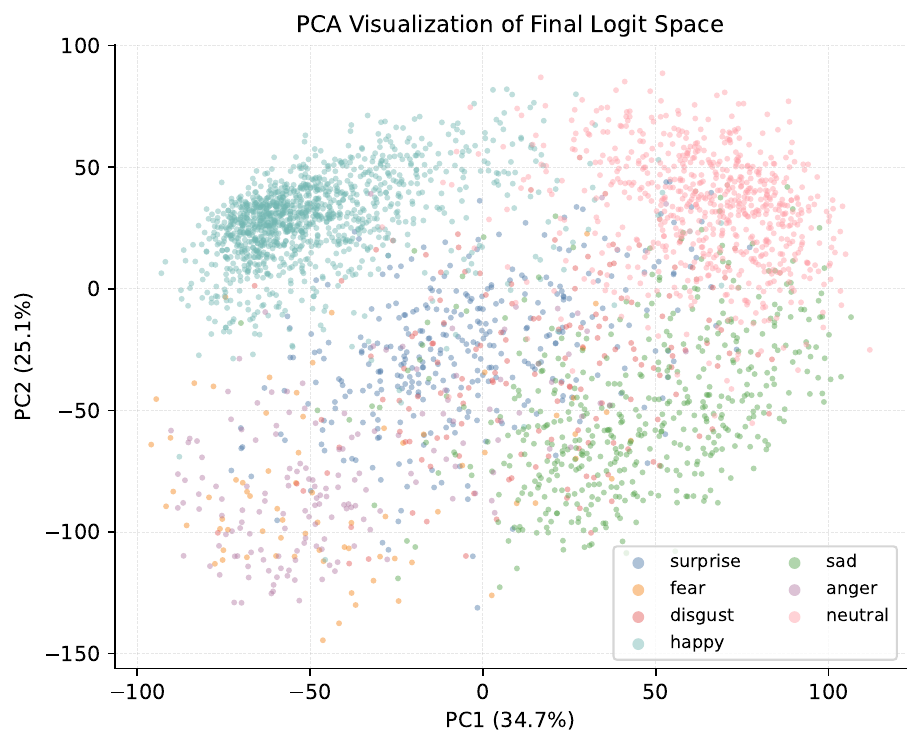}
\par\smallskip\textbf{(b)} Final logit-space PCA
\end{minipage}
\caption{Complementary RAF-DB diagnostics. (a) Input, CLS-only attention, SRE-FER attention, and AU prior. (b) PCA projection of final logits, showing both separated clusters and residual class overlap.}
\label{fig:diagnostics}
\end{figure}

Figure~\ref{fig:diagnostics} complements these class-level statistics. Panel~(a) compares Stage~A (CLS-only DINOv3) with Stage~F (SRE-FER Deploy.); in the shown examples, responses shift from background or expression-irrelevant areas toward AU-related regions. AU mass is $\sum_i A_iM_i$, where $A_i$ is normalized patch attention and $M_i$ is the AU prior mask. In the principal component analysis (PCA) view of panel~(b), happy and neutral form relatively coherent clusters, whereas several boundary-sensitive classes still overlap. The first two components explain 59.8\% of the variance, so the projection is descriptive rather than a quantitative separability measure.

The PCA geometry mirrors the residual confusion matrices: compact clusters coexist with diffuse, mixed bands. Thus SRE-FER recovers part of the local information underused by global aggregation rather than fully disentangling visually similar expressions.

\begin{table}
\caption{Fine-grained F1 results on RAF-DB. Minority and boundary F1 are unweighted means of the class-wise F1 values for \{\textit{fear}, \textit{disgust}\} and \{\textit{fear}, \textit{disgust}, \textit{sad}, \textit{surprise}\}, respectively.}\label{tab:fine_grained}
\centering
\small
\begin{tabular}{lcccc}
\toprule
\textbf{Variant} & \textbf{Acc (\%)} & \textbf{Minority F1 (\%)} & \textbf{Boundary F1 (\%)} & \textbf{Boundary gain (pt.)} \\
\midrule
CLS only & 90.25 & 70.5 & 79.9 & -- \\
AU-guided & 90.88 & 73.0 & 81.1 & +1.2 \\
SRE-FER (Deploy.) & 92.47 & \textbf{75.2} & \textbf{83.5} & \textbf{+3.6} \\
\bottomrule
\end{tabular}
\end{table}

To separate useful corrections from regressions, Table~\ref{tab:repair} compares both settings against the same 2823/3068-correct pre-routing predictions (a Deploy.-level checkpoint evaluated without EMA/SWA smoothing, $\approx$92.01\%). A \emph{fixed} sample changes from incorrect to correct, whereas a \emph{broken} sample changes in the opposite direction.

\begin{table}[!htbp]
\caption{Error repair on RAF-DB relative to the same pre-routing predictions.}\label{tab:repair}
\centering
\footnotesize
\setlength{\tabcolsep}{7pt}
\begin{tabular}{lrrrr}
\toprule
\textbf{Setting} & \textbf{Fixed} & \textbf{Broken} & \textbf{$\Delta$} & \textbf{Acc. (\%)} \\
\midrule
SRE-FER (Deploy.) & 20 & 6 & +14 & 92.47 \\
SRE-FER (Full) & \textbf{32} & 9 & \textbf{+23} & \textbf{92.76} \\
\bottomrule
\end{tabular}
\end{table}

\FloatBarrier

\subsection{Deployment Setting and Recipe Sensitivity}\label{sec:stability}

Table~\ref{tab:seed} complements, rather than replicates, the validation-selected no-TTA result in Table~\ref{tab:comparison}. Its three independent runs use fixed SWA/EMA schedules and optional horizontal-flip TTA, yielding $92.00 \pm 0.15\%$ accuracy and Macro-F1 $0.8677 \pm 0.0020$. Because checkpoint variants and inference protocols differ, this mean measures recipe sensitivity and is not a confidence interval for 92.47\%.

Deploy. discards AU masks and landmarks after training and coordinates only four regional tokens; with $D=1024$ and seven classes, RERA contains 8.54M trainable parameters. Full reuses this readout and additionally selects $B=32$ FPS bases, solves a $32\times32$ ridge system, and pools the top 16 residual tokens per sample. The selector itself is parameter-free, but final server latency and floating-point operations (FLOPs) were not measured.

\begin{table}[!htb]
\caption{Additional seed and checkpoint sensitivity results on RAF-DB.}\label{tab:seed}
\centering
\small
\begin{tabular}{lcccc}
\toprule
\textbf{Seed} & \textbf{Raw Acc (\%)} & \textbf{Selected Variant} & \textbf{Selected Acc (\%)} & \textbf{Macro-F1 (0--1)} \\
\midrule
1 & 91.04 & SWA + hflip TTA & 91.92 & 0.8699 \\
2 & 91.53 & SWA + hflip TTA & 92.18 & 0.8671 \\
3 & 91.36 & EMA + hflip TTA & 91.92 & 0.8661 \\
\midrule
Mean $\pm$ std & $91.31 \pm 0.25$ & -- & $92.00 \pm 0.15$ & $0.8677 \pm 0.0020$ \\
\bottomrule
\end{tabular}
\end{table}

\FloatBarrier

\section{Analysis and Discussion}

The progressive DINOv3 ablation assigns the largest structural change to the complete RERA unit (+0.92), while grouped F1 and confusion matrices place much of the improvement in boundary-sensitive classes. Attention examples indicate where the model responds, and PCA shows the remaining overlap. Unlike POSTER's landmark stream~\cite{ref_poster}, Deploy. reads regional cues directly from DINOv3 tokens, extending the earlier CLS-only test~\cite{ref_mdnre} to dense tokens and in-the-wild benchmarks.

The evidence supports RERA as a unit rather than every internal choice in isolation. AU guidance provides class-conditioned anatomical structure, RERA aggregates and coordinates four coarse regions, and FPS routing probes sample-specific residual tokens. The positive fix/break balance shows that the combined evidence corrects more RAF-DB predictions than it disrupts, but it does not establish that zero initialization or the four-way partition is individually optimal.

The two inference settings serve different roles. Deploy. removes AU and landmark processing and retains the core regional residual readout. Full adds conventional FPS selection and ridge reconstruction, yielding modest 0.19--0.29 point increments across the three datasets and a larger net repair count on RAF-DB.

The current evidence has clear boundaries. Fixed quadrants assume roughly aligned faces and may degrade under large pose or occlusion; the AU probe also shows class-dependent detection failures. Table~\ref{tab:ablation} does not include capacity-matched non-residual heads or LoRA/VPT under the same backbone, and the heatmaps remain qualitative. Adaptive regions, matched readout baselines, cross-domain evaluation, and explicit Full-path profiling are therefore the most direct next steps.

\section{Conclusion}

SRE-FER retains dense DINOv3 patch tokens and adds regional residual correction under training-time AU guidance. On three benchmarks, SRE-FER attains 92.76\% on RAF-DB, 91.32\% on FERPlus, and 67.78\% on AffectNet-7, demonstrating highly competitive performance compared to existing FER methods. Progressive ablations support RERA as a complete readout unit, while PCA exposes residual overlap. The fixed regional partition, unmatched alternative readouts, and unprofiled Full path remain limitations.

%
%
\clearpage
\bibliographystyle{spiebib}
\bibliography{references}

\end{document}